\documentclass[10pt,a4paper,twocolumn]{article}
\usepackage[latin1]{inputenc}
\usepackage[english]{babel}
\usepackage{amsmath}
\usepackage{amsfonts}
\usepackage{amssymb}
\usepackage{graphicx}
\usepackage[left=2cm,right=2cm,top=2cm,bottom=2cm]{geometry}

\usepackage{amssymb}
\usepackage{latexsym}

\usepackage{url}
\usepackage{xcolor}
\definecolor{newcolor}{rgb}{.8,.349,.1}

\usepackage{times}
\usepackage{epsfig}

\usepackage{graphicx}
\usepackage{amsmath}

\usepackage{caption}
\usepackage{makecell}
\usepackage{booktabs}
\usepackage{subfig}
\usepackage{multirow}

\usepackage{multicol}
\usepackage{enumitem}

\usepackage{url}

\usepackage{breakurl}
\usepackage[breaklinks]{hyperref}

\usepackage{algorithm}
\usepackage[noend]{algpseudocode}

\newcommand\Mark[1]{\textsuperscript#1}

\title{Post-comparison mitigation of demographic bias in face recognition using fair score normalization}

\author{Philipp~Terh\"{o}rst\Mark{1}\Mark{2}, Jan Niklas Kolf \Mark{2}, Naser Damer\Mark{1}, Florian Kirchbuchner\Mark{1}\Mark{2}, Arjan Kuijper\Mark{1}\Mark{2}\\
\Mark{1}Fraunhofer Institute for Computer Graphics Research IGD, Darmstadt, Germany\\
\Mark{2}Technical University of Darmstadt, Darmstadt, Germany\\
Email:{\{philipp.terhoerst, naser.damer, florian.kirchbuchner, arjan.kuijper\}@igd.fraunhofer.de}
}

\date{}

\begin{document}

\maketitle

\begin{abstract}

Current face recognition systems achieve high progress on several benchmark tests.
Despite this progress, recent works showed that these systems are strongly biased against demographic sub-groups.
Consequently, an easily integrable solution is needed to reduce the discriminatory effect of these biased systems.
Previous work mainly focused on learning less biased face representations, which comes at the cost of a strongly degraded overall recognition performance.
In this work, we propose a novel unsupervised fair score normalization approach that is specifically designed to reduce the effect of bias in face recognition and subsequently lead to a significant overall performance boost.
Our hypothesis is built on the notation of individual fairness by designing a normalization approach that leads to treating "similar" individuals "similarly".
Experiments were conducted on three publicly available datasets captured under controlled and in-the-wild circumstances.
Results demonstrate that our solution reduces demographic biases, e.g. by up to 82.7\% in the case when gender is considered.
Moreover, it mitigates the bias more consistently than existing works.
In contrast to previous works, our fair normalization approach enhances the overall performance by up to 53.2\% at false match rate of $10^{-3}$ and up to 82.9\% at a false match rate of $10^{-5}$.
Additionally, it is easily integrable into existing recognition systems and not limited to face biometrics.

\end{abstract}

\section{Introduction}
\label{sec:Introduction}











Large-scale face recognition systems are spreading worldwide and are increasingly involved in critical decision-making processes, such as in forensics and law enforcement.
Consequently, these systems also have a growing effect on everybody's daily life.
However, current biometric solutions are mainly optimized for maximum recognition accuracy  \cite{Jain201650YO} and are heavily biased for certain demographic groups \cite{faceAccurate,DBLP:journals/corr/abs-1809-02169,Furl2002FaceRA,Phillips:2011:OEF:1870076.1870082,pmlr-v81-buolamwini18a, garvie2016perpetual}.
This means that, for example, specific demographic groups can be falsely identified as black-listed individuals more frequently than other groups.
Consequently, there is an increased need that guarantees fairness for biometric solutions \cite{pmlr-v81-buolamwini18a,goodman2016regulations,pmlr-v28-zemel13} to prevent discriminatory decisions.

From a political perspective, there are several regulations to guarantee fairness.
Article 7 of the Universal Declaration on Human Rights and Article 14 of the European Convention of Human Rights ensure people the right to non-discrimination.
Also the General Data Protection Regulation (GDPR) \cite{Voigt:2017:EGD:3152676} aims at preventing discriminatory effects (article 71).
In spite of these political efforts, several works \cite{Phillips:2011:OEF:1870076.1870082,pmlr-v81-buolamwini18a,faceAccurate,DBLP:journals/corr/abs-1809-02169,Furl2002FaceRA, garvie2016perpetual} showed that commercial \cite{pmlr-v81-buolamwini18a}, as well as open-source \cite{faceAccurate} face recognition systems, are strongly biased towards different demographic groups.
Consequently, there is an increased need for fair and unbiased biometric solutions \cite{faceAccurate, garvie2016perpetual}.

Recent works mainly focused on learning less-biased face representations \cite{gong2019debface,Liang_2019_CVPR, wang2019mitigate,DBLP:journals/corr/abs-1806-00194, WANG201935,Kortylewski_2019_CVPR_Workshops,DBLP:conf/cvpr/00010S0C19} for specific demographics.
However, this requires computationally expensive template-replacement of the whole database if the recognition system is updated.
Moreover, the bias-mitigation often comes at the cost of a strong decrease in recognition performance.

In this work, we propose a novel and unsupervised fair score normalization bias mitigation approach that is easily-integrable.
Unlike previous work, increasing fairness also leads to an improved performance of the system in total.
Our theoretical motivation is based on the notation of individual fairness \cite{Dwork:2012:FTA:2090236.2090255}, resulting in a solution that treats similar individuals similarly and thus, more fairly.
The proposed approach clusters samples in the embedding space such that similar identities are categorized without the need for pre-defined demographic classes.
For each cluster, an optimal local threshold is computed and used to develop a score normalization approach that ensures a more individual, unbiased, and fair treatment.
The experiments are conducted on three publicly available datasets captured under controlled and in-the-wild conditions and on two face embeddings.
To justify the concept of our fair normalization approach, we provide a visual illustration that demonstrates 
(a) the suitability of the notation of individual fairness for face recognition and (b) the need for more individualized treatment of face recognition systems.
The results show a higher consistency and efficiency of our unsupervised normalization approach compared to related works.
It efficiently mitigates demographic-bias by up to 82.7\% while consistently enhancing the total recognition performance by up to 82.9\%.
The source code for this work is available at the following link\footnote{\url{https://github.com/pterhoer/FairScoreNormalization}}.

\section{Related work}
\label{sec:RelatedWork}

Bias in biometrics was found in several disciplines such as presentation attack detection \cite{DBLP:journals/corr/abs-2003-03151}, biometric image quality estimation \cite{DBLP:journals/corr/abs-2004-01019}, and the estimation of facial characteristics \cite{DBLP:conf/fusion/TerhorstD0K19,DBLP:conf/btas/Terhoerst19,DBLP:conf/eccv/DasDB18}.
In face biometrics, bias might be induced by non-equally distributed classes in training data \cite{Kortylewski_2019_CVPR_Workshops, DBLP:journals/corr/abs-1806-00194}.
This results in face recognition performances that are strongly influenced by demographic attributes \cite{FRVT2019}.
These findings motivated the recent research towards mitigating bias in face recognition solutions.
Recent works focused on learning less-biased face representations through adversarial learning \cite{gong2019debface,Liang_2019_CVPR}, margin-based approaches \cite{wang2019mitigate,DBLP:journals/corr/abs-1806-00194}, or data augmentation \cite{WANG201935,Kortylewski_2019_CVPR_Workshops,DBLP:conf/cvpr/00010S0C19}.


In \cite{gong2019debface}, Gong, Liu, and Jain proposed de-biasing adversarial network.
This network consists of one identity classifier and three demographic classifiers (gender, age, race) and aims at learning disentangled feature representations for unbiased face recognition.
Liang et al. \cite{Liang_2019_CVPR} proposed a two-stage method for adversarial bias mitigation.
First, they learn disentangled representations by a one-vs-rest mechanism and second, they enhance the disentanglement by additive adversarial learning.

Also margin-based approaches were proposed to reduce bias in face recognition systems.
In \cite{wang2019mitigate}, Wang et al. applied reinforcement learning to determine a margin that minimizes ethnic bias.
Huang et al. \cite{DBLP:journals/corr/abs-1806-00194} proposed a cluster-based large-margin local embedding approach to reduce the effect of local data imbalance and thus, aims at reducing bias coming from unbalanced training data.

Finally, data augmentation methods were presented for fairer face recognition.
In \cite{WANG201935}, Wang et al. proposed a large margin feature augmentation to balance class distributions.
Kortylewski et al. \cite{Kortylewski_2019_CVPR_Workshops} proposed a data augmentation approach with synthetic data generation and Yin et al. \cite{DBLP:conf/cvpr/00010S0C19} proposed a center-based feature transfer framework to augment under-represented samples.

So far, previous work mainly focused on learning less-biased face representations.
However, updating the recognition system with one of these approaches will require a computationally expensive template-replacement of the whole database. Moreover, it requires that a face image for every enrolled individual is additionally stored.
Therefore, more integrable solutions were proposed  \cite{DBLP:conf/iwbf/TerhorstD0K20, Srinivas_2019_CVPR_Workshops}.
In \cite{DBLP:conf/iwbf/TerhorstD0K20}, a fair template comparison approach was proposed to mitigate ethnic-bias at the decision-level.
While this work provides an easily-integrable solution, it requires parameter-tuning towards the biased attribute and degrades the total recognition performance.
In \cite{Srinivas_2019_CVPR_Workshops}, the combination of multiple face recognition systems was investigated to mitigate bias.
Their best approach shows an improvement of the total recognition performance but at the cost of only marginal and unstable bias reductions, since this method was originally not developed for mitigating bias.
Consequently, we compare our fair normalization approach against these two solutions.

A recent work \cite{DBLP:journals/corr/abs-2009-09918} showed that many soft-biometric attributes (beyond identity-related information) are encoded in face templates that might lead to differences in the recognition performance (bias).
Consequently, more unsupervised solutions on bias-mitigation are needed to address these concerns.

To the best of our knowledge, this is the first score-normalization approach that is specifically designed for bias-mitigation in face recognition systems \cite{9086771}.
In contrast to previous works, our method jointly (a) does not need additional soft-biometric labels during training or inference time, (b) can be easily integrated into existing face recognition systems, (c) enhances the total face recognition performance and (d) leads to a consistent bias-mitigation.

\section{Methodology}
\label{sec:Methodology}







The goal of this work is to enhance the fairness of existing face recognition systems in an easily-integrable manner.
In this work, we follow the notation of individual fairness \cite{Dwork:2012:FTA:2090236.2090255}.
This notation emphasizes that similar individuals should be treated similarly.
We transfer this idea to the embedding and score level to propose a novel fair group-based score normalization method, without the need for pre-defined demographic groups.
The proposed approach is able to treat all identities more individually and therefore, increase the group-related, as well as the total, recognition performance.

\vspace{-3mm}
\subsection{Fair group score normalization}
Our proposed solution is presented assuming a set of face embeddings $X = \left( X_{train} \cup X_{test} \right)$ with the corresponding identity information $y = \left( y_{train} \cup y_{test} \right)$, both partitioned into test and training set.
\paragraph*{Training phase}
During training phase, a k-means cluster algorithm \cite{10.2307/2346830} is trained on $X_{train}$ to split the embedding space into $k$ clusters ($k=100$ in our experiment).
For each cluster $c \in \left\lbrace 1,\dots, k \right\rbrace$, an optimal 
threshold for a false match rate of $10^{-3}$
is computed using the genuine and imposter scores of cluster $c$
\begin{align}
gen_c &= \left\lbrace s_{ij} \, | \, \textit{ID}(i) = \textit{ID}(j),\,  i \neq j, \, \forall \, i  \in C_c,  \right\rbrace \label{eq:GenScores} \\
imp_c &= \left\lbrace s_{ij} \, | \, \textit{ID}(i) \neq \textit{ID}(j),\,  \forall \, i  \in C_c,  \right\rbrace .
\end{align}
The genuine score set $gen_c$ of cluster $c$ includes the all comparison scores of samples $i$ and $j$ that come from the same identity $(\textit{ID}(i) = \textit{ID}(j))$, where at least one sample lies within cluster $c$ $(i \in C_c)$.
Conversely, the imposter score set $imp_c$ of cluster $c$ is defined as all comparison scores $s_{ij}$ from different identity pairs $(\textit{ID}(i) \neq \textit{ID}(j))$ where at least one sample lies within cluster $c$ $(i \in C_c)$.
The local threshold for each cluster $c$ is denoted as $thr(c)$.
Furthermore, the threshold for the whole training set $X_{train}$ is calculated and denoted as the global threshold $thr_G$.


\paragraph*{Operation phase}

During operation phase, the normalized comparison score $\hat{s}_{ij}$ should be computed to determine if sample $i$ and $j$ belong to the same identity.
Firstly, the corresponding clusters for both samples are computed. 
The cluster thresholds for sample $i$ and $j$ are denoted as $thr_i$ and $thr_j$.
Secondly, these cluster thresholds, as well as the global threshold $thr_G$, are used to calculate the normalized score
\begin{align}
&\hat{s}_{ij} = s_{ij} - \frac{1}{2} \left( \Delta thr_i + \Delta thr_j  \right),
\end{align}
where
\begin{align}
\Delta thr_i = thr_i - thr_G,
\end{align} 
describes the local-global threshold difference for sample $i$.

\subsection{Discussion}
The goal of this score normalization approach is to introduce individual fairness in a biometric system and thus, reduce the discriminatory behavior of face recognition systems.
The notation of individual fairness emphasizes that similar individuals should be treated similarly.
We incorporate this statement in our normalization method using clustering and local thresholds.
Clustering in the embedding space identifies similar individuals and local cluster thresholds enable approximately individual treatment.

The choice of the individuality parameter $k$ defines the number of clusters for our fair score normalization and is crucial for the recognition performance.
A small $k$ (e.g. $k=2$) refers to a less individual normalization of the score, while a very large $k$ reduces the number of samples per clusters and thus, the quality of the local thresholds.

To further improve the performance of the fair score normalization approach, it can be considered to fine-tune the utilized face recognition system with real-world data.
However, we did not follow that idea in this work to keep the focus of our experiments on generalizability.
Moreover, if annotated data is available the clustering process can be replaced by forming clusters based on the attribute annotations.
This will lead to more specific clusters and thus, the similarity definition can be adapted to the problem.
However, this will turn our unsupervised solution into a supervised approach that is only able to reduce the bias of pre-defined attributes.

\vspace{-3mm}
\subsection{How does fair normalization affect different sample pairs?}
In the following, we discuss how the proposed fair normalization approach affects biased and unbiased genuine and imposter pair comparisons.
With samples from a biased group, we refer to a group of samples that achieve significantly weaker face recognition performances than other groups of individuals.
With our methodology, these (biased) groups are characterized by low local thresholds.

\textit{Biased genuine pair - }
Assuming that an identity $\mathcal{I}$, with samples $i$ and $j$, belongs to a \textit{biased} group, their comparison score $s_{ij}=0.4$ will be low.
Since this is lower than the global threshold $thr_G=0.6$, the decision for this genuine pair will be falsely made towards imposter.
Since these samples belong to a biased cluster, the recognition performance within is low and so are the local thresholds $thr_i=thr_j=0.3$.
The low local thresholds lead to a negative local-global threshold difference $\Delta thr_i = \Delta thr_j = 0.3-0.6=-0.3$ and thus, the normalized comparison score $\hat{s}_{ij}=0.4-\frac{1}{2}(-0.3-0.3)=0.7$ increases.
Since $\hat{s}=0.7>thr_G=0.6$, the system now comes to the correct genuine decision with the proposed normalization method.

\textit{Unbiased genuine pair - }
Assuming that an identity $\mathcal{I}$, with samples $i$ and $j$, belongs to an \textit{unbiased} group, their comparison score $s_{ij}=0.9$ will be high.
Since these samples belong to an unbiased cluster, the performance within is high and so are the local thresholds $thr_i=thr_j=0.7$.
The low local thresholds leads to a positive $\Delta thr_i = \Delta thr_j = 0.9-0.6=+0.2$ and thus, the normalized comparison score $\hat{s}_{ij}=0.9-\frac{1}{2}(0.2+0.2)=0.7$ increases.
Since $\hat{s}=0.7>thr_G=0.6$, the system still come to the correct genuine decision.

\textit{Imposter pair - }
For imposter pairs $(i,j)$, three situations have to be considered depending cluster-correspondence of the two samples.
The first one refers to the case in which one of the two samples belong to a cluster with a low local threshold, while the other belongs to one with a large local threshold.
In this case, our normalization approach is only marginally changing the comparison score.
Therefore, the verification decision is unchanged.

In the second case, both samples belong to clusters with high local thresholds.
Consequently, the score is highly reduced and thus, the probability for a false match decreases.

The third case is the most critical, where both samples belong to clusters with low local thresholds.
If both samples belong to different clusters, then their embeddings are dissimilar and will result in a low comparison score.
Consequently, the risk of a false match is low.
If both samples belong to the same cluster, their embeddings are similar and thus, there is a high risk of a false match.
However, our method is especially optimized for exactly this (critical) case, since the local thresholds are computed based on intra-cluster performance.
Consequently, the false acceptance rate with our normalization is lower or equal than the unnormalized case.

\textit{Main error - }
The main error that can appear with the normalization approach happens at the border of two adjacent clusters with high differences in the local thresholds.
Comparing similar samples at the border of these clusters might lead to overcorrections of the scores. 
However, Figure \ref{fig:t-SNE} showed that this is rarely the case.
Moreover, this can be prevented by a sufficient choice of $k$, since $k$ determines the number of clusters and a larger number of clusters lead to more fine-grained local thresholds of adjacent clusters.

\vspace{-3mm}
\section{Experimental setup} 
\label{sec:ExperimentalSetup}

\textit{Database - }
In order to evaluate the face recognition performance of our approach under controlled and unconstrained conditions, we conducted experiments on the public available Adience \cite{Eidinger:2014:AGE:2771306.2772049}, ColorFeret \cite{ColorFERET}, and Morph \cite{DBLP:conf/fgr/RicanekT06} datasets.
ColorFeret \cite{ColorFERET} consists of 14,126 images from 1,199 different individuals with different poses under controlled conditions.
Furthermore, a variety of face poses, facial expressions, and lighting conditions are included in the dataset.
The Adience dataset \cite{Eidinger:2014:AGE:2771306.2772049} consists of over 26.5k images from over 2.2k different subjects under unconstrained imaging conditions.
Morph \cite{DBLP:conf/fgr/RicanekT06} contains 55,134 images from 13,618 subjects.
The ages range from 16 to 77 with a median of 33 years.
While Adience contains additional information about gender and age, ColorFeret and Morph also provide labels regarding the subject's ethnicities.
The distribution of these attributes in the databases is shown in Table \ref{tab:CV_Data}.
In the experiments, this information is used to investigate the face recognition performance for several demographic groups.

\begin{table}[h]
\renewcommand{\arraystretch}{1.0}
\centering
\caption{Attribute distribution of the images in the used databases.}
\label{tab:CV_Data}
\begin{tabular}{llrrr}
\Xhline{2\arrayrulewidth} 
          &               & \multicolumn{3}{c}{Database}  \\
          \cmidrule(lr){3-5}
Attribute & Class         & Adience & ColorFeret & Morph  \\
\hline
Gender    & Male          & 51.6\%   & 64.6\%     & 84.6\% \\
          & Female        & 48.4\%   & 35.4\%     & 15.4\% \\
          \hline
Age       & \textless{}20 & 40.0\%   & 1.4\%      & 17.2\% \\
          & 20-30         & 33.2\%  & 34.9\%     & 28.1\% \\
          & 30-40         & 15.9\%   & 27.9\%      & 28.3\% \\
          & 40+           & 10.9\%   & 35.8\%     & 26.4\% \\
          \hline
Ethnicity & Asian         & -       & 23.3\%     & 0.4\%  \\
          & Black         & -       & 7.6\%     & 77.2\% \\
          & White         & -       & 62.6\%     & 19.2\% \\
          & Other         & -       & 6.5\%     & 3.2\%  \\
          \hline
          \Xhline{2\arrayrulewidth} 
\end{tabular}
\end{table}


\textit{Evaluation metrics - }
In this work, we will report the recognition performances in terms of false non-match rate (FNMR) at fixed false match rates (FMR).
As recommended by the European Border Guard Agency Frontex \cite{FrontexBestPractice}, we will use FMR thresholds of $10^{-3}$ and smaller.
To evaluate the amount of demographic bias in the recognition performance, the recognition performance is evaluated within all subgroups and the standard deviation (STD) of these group-specific performances is reported.
A low STD refers to a more unbiased attribute performance since the performances of the different attribute classes are similar.
In contrast, a high STD refers to a biased attribute performance with strong performance differences between the attribute classes.


\textit{Workflow details - }
For the comparison of two samples, both face images get aligned, scaled, and cropped. 
Then, the preprocessed images are passed into a face recognition model resulting in a face template for each image.
The comparison of two embeddings is done using cosine-similarity.
In this work, we use FaceNet\footnote{\url{https://github.com/davidsandberg/facenet} \cite{DBLP:journals/corr/SchroffKP15} and VGGFace\footnote{\url{https://github.com/ox-vgg/vgg_face2}} \cite{DBLP:conf/bmvc/ParkhiVZ15}.
Both models were trained on MS-Celeb-1M \cite{DBLP:journals/corr/GuoZHHG16}.
Moreover, the preprocessing for FaceNet was done based on \cite{King:2009:DML:1577069.1755843} and for VGGFace, the preprocessing follows the methodology described in \cite{7553523}.
}.
The clustering needed for our fair score normalization is calculated on the same features that are used for the verification.
%
For all experiment scenarios, subject-disjoint 5-fold cross-validation is utilized
and in each iteration, all possible positive and negative face combinations pairs are evaluated.

\textit{Baseline approaches - }
We evaluate our fair score normalization approach in comparison with two bias-mitigating works \cite{DBLP:conf/iwbf/TerhorstD0K20, Srinivas_2019_CVPR_Workshops} that, just as our solution, act beyond template-generation and thus, are easily-integrable as well.
In \cite{DBLP:conf/iwbf/TerhorstD0K20}, a fair template comparison (FTC) approach is proposed aiming at mitigating ethnic-bias.
For our experiments, we trained the model with $\lambda=0.5$.
This choice is based on the recommendation of \cite{DBLP:conf/iwbf/TerhorstD0K20}.
In \cite{Srinivas_2019_CVPR_Workshops}, base normalization and score-level fusion (SLF) strategies are investigated for mitigating bias in face recognition systems.
We use their best working approach, namely min-max normalization with a simple sum-fusion rule, as an additional baseline in our experiments combining both utilized face embeddings.

\textit{Investigations - }
The investigations of this work are divided into four parts:
\begin{enumerate}
\setlength{\itemsep}{0pt}%
\item We first visually demonstrate the need for a more individual treatment in face recognition systems. Moreover, we show that our approach is able to treat similar individuals more similarly.
\item We investigate the effect of the individuality parameter $k$ over a wide parameter range since this critically affects the effectiveness of the proposed approach.
\item We investigate the bias-mitigation performance of the unmodified baseline, our normalization approach, and state-of-the-art approaches. Therefore, the intra-class verification performance is investigated for different demographic attributes and the attribute-bias is measured and compared.
\item We finally investigate the overall verification performance to prove that, unlike previous works, our approach enhances the overall performance while mitigating demographic-bias.
\end{enumerate}

\section{Results}
\label{sec:Results}

\subsection{Visual demonstration of the need for individuality}

\begin{figure*}[h]
\centering
\includegraphics[width=0.7\textwidth]{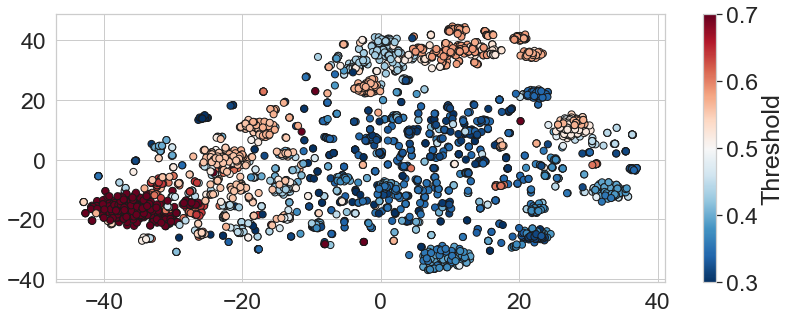}
\caption{Visualizations of the Adience FaceNet embeddings using t-SNE \cite{vanDerMaaten2008}. Each individual is represented as a point and each point is colored based on its optimal local threshold $thr_k$. 
The formation of several clusters with similar local thresholds shows that our approach is able to identify similar individuals and to treat them similarity. The large variation of optimal local thresholds (0.3-0.7) demonstrates the need of this more individual, and thus fair, treatment.}
\label{fig:t-SNE}
\end{figure*}

Since our approach is based on the idea of individual fairness, we first want to visually demonstrate why this notation is suitable for face recognition.
Figure \ref{fig:t-SNE} shows an t-SNE visualization of the embedding space for the dataset Adience.
The t-SNE algorithm maps the high-dimensional embedding space into a two-dimensional space such that similar samples in the high-dimension space lie closely together in two dimensions.
Furthermore, each sample is colored based on the local thresholds computed by the proposed approach.
Two observations can be made from this figure:
first, it shows that there are several clusters with similar local thresholds in the embedding space.
Consequently, our proposed approach is able to identify similar identities and to treat them similarly (through similar local thresholds). 
Second, it shows that the optimal thresholds for each cluster vary significantly from 0.3 to 0.7.
This widespread of optimal local thresholds demonstrates the need for a more individual, and thus fair, treatment.

\vspace{-2mm}
\subsection{The choice of the individuality parameter $k$}
\label{sec:ResultsIndividualityAnalysis}

\begin{figure}[h]
\centering       
  \subfloat[Adience - FaceNet]{%
       \includegraphics[width=0.20\textwidth]{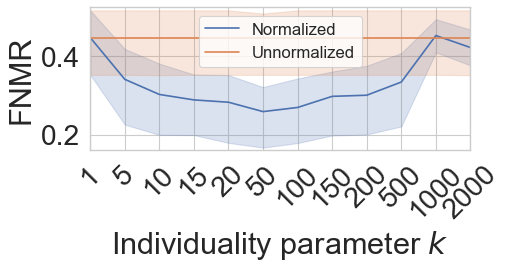}}
  \subfloat[Adience - VGGFace]{%
       \includegraphics[width=0.20\textwidth]{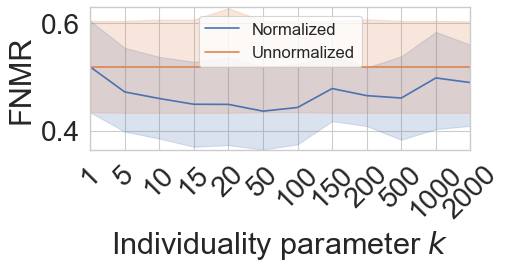}} 
         
  \subfloat[ColorFeret - FaceNet]{%
       \includegraphics[width=0.20\textwidth]{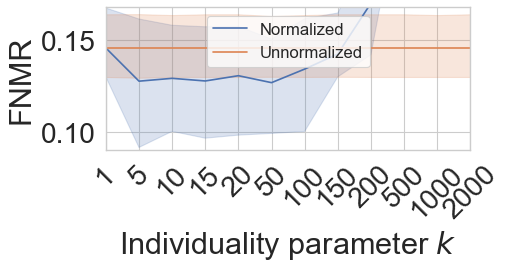}}
  \subfloat[ColorFeret - VGGFace \label{fig:Individuality_VGG_FaceNet}]{%
       \includegraphics[width=0.20\textwidth]{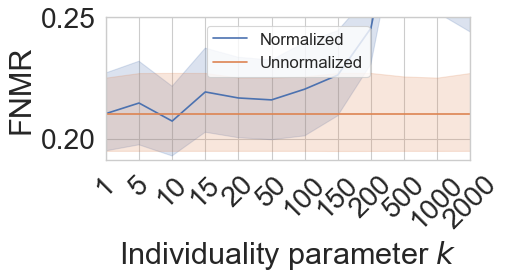}} 
       
         \subfloat[Morph - FaceNet]{%
       \includegraphics[width=0.20\textwidth]{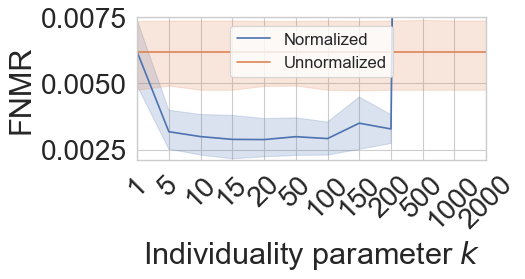}}
  \subfloat[Morph - VGGFace]{%
       \includegraphics[width=0.20\textwidth]{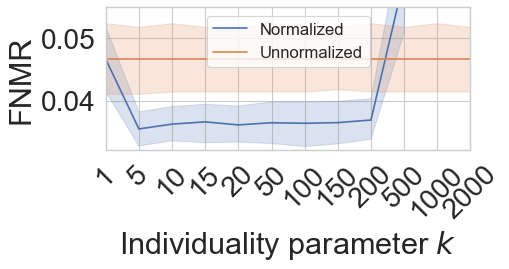}} 
\caption{Analysis of the verification performance at a FMR of $10^{-3}$ based on the individuality parameters $k$. The proposed normalization approach (blue) is compared against the unnormalized baseline (orange). The analysis includes three datasets and two face embeddings. The shaded areas represent the standard deviation over the 5 cross-validation folds. Individuality parameters around $k \approx 100$ show a generally a stable performance improvement.}
\label{fig:IndividualityAnalysis}
\end{figure}

In this section, we analyse the sensitivity of the individuality parameter $k$ and justify our choice for $k=100$.
Figure \ref{fig:IndividualityAnalysis} shows verification performances of the proposed fair normalization over different individuality parameters $k$ on all datasets and both face embeddings.
Moreover, the unnormalized baseline is shown.
For $k=1$, the normalization does not change the scores and thus, the same performance is observed with and without the normalization.
For $k\geq 1$, the verification performance increases, since our fair score normalization approach leads to more individual treatment of each sample.
This can be observed in all cases (a,b,c,e,f), except for Figure \ref{fig:Individuality_VGG_FaceNet}. 
In this case, the clustering algorithm produces clusters of unequal sizes leading to performance degradation.
However, this still lies within the standard deviation of the unnormalized case.
Moreover, this still leads to a strong bias-mitigation, as we will see in Section \ref{sec:ResultsBiasAnalysis}.
If $k$ is large, the number of samples per cluster decreases.
Since these are necessary to determine the local thresholds, the quality of these decreases.
This leads to unreliable thresholds and thus, inaccurate recognition performances.
For all datasets and both embeddings, this performance drop can be observed for large $k$.
However, individuality parameters around $k \approx 100$ show a generally stable performance.
Therefore, we choose $k=100$ for our experiments.

\subsection{Analysis of the demographic-bias}

\label{sec:ResultsBiasAnalysis}

\begin{table*}[h]
\setlength{\tabcolsep}{7pt}
\renewcommand{\arraystretch}{1.0}
\centering
\caption{Intra-class recognition performance of our approach: the performance is shown in terms of FNMR@$10^{-3}$ FMR for FaceNet and VGGFace embeddings. The unnormalised (unnorm.) and normalized (norm.) performance within each attribute class is reported with the corresponding performance change.
In many cases, the proposed normalization approach enhances the fairness by strongly improving the performance of under-performing classes.
In other cases, the approach leads to a performance adaptation to minimize the performance differences between the groups, leading to more fair recognition decisions as shown in Table \ref{tab:BiasReductionPerformance}.
}
\label{tab:SubgroupPerformance}
\begin{tabular}{lllrrrrrr}
\Xhline{2\arrayrulewidth} 
           &           &               & \multicolumn{3}{c}{FaceNet}                        & \multicolumn{3}{c}{VGGFace}                        \\
           \cmidrule(lr){4-6} \cmidrule(lr){7-9}
Database   & Attribute & Class         & Unnorm. & Norm. & Perf. change & Unnorm. & Norm. & Perf. change \\
\hline
Adience    & Gender    & Male          & 0.5129       & 0.2600     & 49.3\%                 & 0.4636       & 0.4462     & 3.8\%                  \\
           &           & Female        & 0.3837       & 0.2823     & 26.4\%                 & 0.4703       & 0.4985     & -6.0\%                 \\
           & Age       & 0-2           & 0.7764       & 0.7641     & 1.6\%                  & 0.7861       & 0.7753     & 1.4\%                  \\
           &           & 4-6           & 0.6069       & 0.5838     & 3.8\%                  & 0.7327       & 0.7417     & -1.2\%                 \\
           &           & 8-12          & 0.4327       & 0.3804     & 12.1\%                 & 0.4527       & 0.4769     & -5.3\%                 \\
           &           & 15-20         & 0.7677       & 0.4890     & 36.3\%                 & 0.4358       & 0.4242     & 2.7\%                  \\
           &           & 25-32         & 0.2264       & 0.1540     & 32.0\%                 & 0.3174       & 0.3049     & 3.9\%                  \\
           &           & 38-43         & 0.1766       & 0.1631     & 7.6\%                  & 0.2670       & 0.3002     & -12.4\%                \\
           &           & 48-53         & 0.2253       & 0.1398     & 37.9\%                 & 0.2859       & 0.3018     & -5.6\%                 \\
           &           & 60-100        & 0.1224       & 0.1140     & 6.9\%                  & 0.2468       & 0.2451     & 0.7\%                  \\
           \hline
ColorFeret & Gender    & Male          & 0.1635       & 0.1424     & 12.9\%                 & 0.2252       & 0.2421     & -7.5\%                 \\
           &           & Female        & 0.2167       & 0.1891     & 12.7\%                 & 0.2732       & 0.2704     & 1.0\%                  \\
           & Age       & 10-20         & 0.2118       & 0.1818     & 14.2\%                 & 0.2912       & 0.2873     & 1.3\%                  \\
           &           & 21-30         & 0.1506       & 0.1071     & 28.9\%                 & 0.2059       & 0.2070     & -0.5\%                 \\
           &           & 31-40         & 0.1452       & 0.1459     & -0.5\%                 & 0.1842       & 0.2208     & -19.9\%                \\
           &           & 40+           & 0.0933       & 0.1212     & -29.9\%                & 0.1701       & 0.2034     & -19.6\%                \\
           & Ethnicity & Asian         & 0.3177       & 0.2553     & 19.6\%                 & 0.3099       & 0.3170     & -2.3\%                 \\
           &           & Black         & 0.2489       & 0.2361     & 5.1\%                  & 0.4120       & 0.3736     & 9.3\%                  \\
           &           & White         & 0.1089       & 0.1282     & -17.7\%                & 0.2085       & 0.2228     & -6.9\%                 \\
           &           & Other         & 0.1424       & 0.1417     & 0.5\%                  & 0.2217       & 0.2112     & 4.7\%                  \\
           \hline
Morph & Gender    & Male          & 0.0059 & 0.0031 & 47.5\%  & 0.0463 & 0.0362 & 21.8\% \\
      &           & Female        & 0.0364 & 0.0153 & 58.0\%  & 0.1220  & 0.1062 & 13.0\% \\
      & Age       & \textless{}20 & 0.0056 & 0.0034 & 39.3\%  & 0.0648 & 0.0585 & 9.7\%  \\
      &           & 20-29         & 0.0039 & 0.0019 & 51.3\%  & 0.0461 & 0.0398 & 13.7\% \\
      &           & 30-39         & 0.0081 & 0.0041 & 49.4\%  & 0.0495 & 0.0404 & 18.4\% \\
      &           & 40+           & 0.0137 & 0.0064 & 53.3\%  & 0.0586 & 0.0472 & 19.5\% \\
      & Ethnicity & African       & 0.0037 & 0.0036 & 2.7\%   & 0.0431 & 0.0389 & 9.7\%  \\
      &           & Asian         & 1.0000  & 0.8036 & 19.6\%  & 1.0000  & 1.0000  & 0.0\%  \\
      &           & European      & 0.0069 & 0.0077 & -11.6\% & 0.0888 & 0.086  & 3.2\%  \\
      &           & Hispanic      & 0.0062 & 0.0057 & 8.1\%   & 0.0396 & 0.0431 & -8.8\% \\
           \Xhline{2\arrayrulewidth} 
\end{tabular}
\end{table*}

\begin{table*}[]
\setlength{\tabcolsep}{5pt}
\renewcommand{\arraystretch}{1.0}
\centering
\caption{Analysis of the bias reduction of the proposed approach (Ours) in comparison with two previous works (SLF \cite{Srinivas_2019_CVPR_Workshops} and FTC \cite{DBLP:conf/iwbf/TerhorstD0K20}). The bias is measured in terms of STD of the class-wise FNMRs at a FMR of $10^{-3}$.
Unlike both previous works, our proposed approach mitigates bias effectively and consistently.}
\label{tab:BiasReductionPerformance}
\begin{tabular}{llrrrrrrrr}
\Xhline{2\arrayrulewidth} 
           &           & \multicolumn{4}{c}{FaceNet}                       & \multicolumn{4}{c}{VGGFace}                       \\
           \cmidrule(lr){3-6} \cmidrule(lr){7-10}
           &           &     Bias (STD)         & \multicolumn{3}{c}{Bias reduction} &         Bias (STD)     & \multicolumn{3}{c}{Bias reduction} \\
           \cmidrule(lr){4-6} \cmidrule(lr){8-10}
Database   & Attribute & Baseline & SLF  \cite{Srinivas_2019_CVPR_Workshops}  & FTC \cite{DBLP:conf/iwbf/TerhorstD0K20} & Ours   & Baseline & SLF  \cite{Srinivas_2019_CVPR_Workshops} & FTC \cite{DBLP:conf/iwbf/TerhorstD0K20} & Ours     \\
\hline
Adience    & Gender    & 0.0646 & 68.5\%  & -44.9\% & 82.7\% & 0.0262 & -112.7\%  & -135.1\% & -79.2\% \\
           & Age       & 0.2515 & 11.9\%  & 45.9\%  & 8.9\%  & 0.1935 & -0.9\% & 100.0\%  & 2.0\%    \\
           \hline
ColorFeret & Gender    & 0.0266 & -8.4\%  & -85.7\% & 12.2\% & 0.0142 & -21.0\% & -81.3\%  & 41.0\%   \\
           & Age       & 0.0420 & 12.8\%  & -56.6\% & 32.5\% & 0.0339 & -47.0\% & -237.1\% & 27.8\%   \\
           & Ethnicity & 0.0833 & 34.9\%  & 5.9\%   & 32.8\% & 0.0673 & 25.0\%  & 39.3\%   & 17.4\%   \\
           \hline
Morph      & Gender    & 0.0216 & -25.9\% & -18.0\% & 60.0\% & 0.0503 & 49.2\%  & -25.0\%  & 5.8\%    \\
           & Age       & 0.0043 & 4.3\%   & -28.4\% & 56.2\% & 0.0084 & 20.4\%  & -108.7\% & 1.6\%    \\
           & Ethnicity & 0.4972 & 0.4\%   & 24.5\%  & 19.8\% & 0.3756 & -31.9\% & 3.8\%    & 20.4\%    \\
           \Xhline{2\arrayrulewidth} 
\end{tabular}
\end{table*}

Our fair normalization approach aims at mitigating biased recognition decisions of unknown origins.
This section analyses this aspect.
In Table \ref{tab:SubgroupPerformance}, the intra-class recognition performance (in terms of FNRM@$10^{-3}$FMR) is shown for several demographic classes with and without our normalization approach.
Table \ref{tab:BiasReductionPerformance} uses this information to measure the attribute-specific bias in the recognition performances and compares it with previous works.
For most attribute classes, the intra-class recognition performance with our fair normalization approach leads to strong enhancements of up to 58\%.
However, for some classes the recognition performance decreases.
This happens when an intra-class recognition performance is much stronger for one class compared to the other classes for this attribute.
Since our fair normalization approach aims at enhancing fairness, and thus reduces the performance differences between the different attribute classes, (a) weak classes have to be improved or (b) strong classes have to be adjusted.
For instance, the second case happens in ColorFeret for age and ethnicity.
The age classes [31-40] and [40+] and white ethnicities perform outstanding well without our normalization and they get adjusted to more closely match the performance of the other attribute classes.

The effectiveness of the proposed normalization approach is shown in Table \ref{tab:BiasReductionPerformance} and compared with previous works.
Here, the bias of an attribute is determined by its standard deviation of the attribute performances.
Moreover, the bias reduction rates are shown.
Positive values indicate a strong bias-mitigation and vice versa. 
Please note that the gender-bias on Adience using VGGFace features is already very low and consequently leads to an increase of gender-specific bias on all investigated approaches.
SLF \cite{Srinivas_2019_CVPR_Workshops} achieves high bias reduction rates in some cases.
However, in 7 out of the 16 cases it even increases the class-biases.
FTC \cite{DBLP:conf/iwbf/TerhorstD0K20} also increases the class-biases in many cases.
Just the ethnic-bias is consistently reduced.
This might relate to the choice of the fairness parameter $\lambda=0.5$ which is recommended in \cite{DBLP:conf/iwbf/TerhorstD0K20} and optimized to mitigation of ethnic-bias.
For our approach, the biases from various origins are consistently mitigated and bias reduction rates of up to 82.7\% are achieved.

\subsection{The global face recognition performance}

This section investigates the overall face recognition performance of our bias-mitigation approach and previous works.
Table \ref{tab:GlobalPerformance} shows the verification performance of FaceNet and VGGFace features on three databases at three decision thresholds.
The performance is reported for the unmodified baseline (Base), for our fair normalisation approach (Ours) and previous works (SLF \cite{Srinivas_2019_CVPR_Workshops} and FTC \cite{DBLP:conf/iwbf/TerhorstD0K20}).
Bias-mitigation often comes at the cost of a decreasing recognition performance.
This can be seen for SLF and FTC.
For example, the overall recognition performance of SLF on FaceNet features decreases in every case on the Morph dataset.
For FTC, the performance decreases in most cases as well.
In contrast, our proposed approach significantly enhances the global recognition performance by up to 82.9\%, while effectively mitigating bias.
Just in one out of 17 cases, the performance slightly decreases due to the failed clustering as discussed in Section \ref{sec:ResultsIndividualityAnalysis}.

\begin{table*}[h]
\renewcommand{\arraystretch}{1.0}
\centering
\setlength{\tabcolsep}{2.7pt}
\caption{Investigation of the overall recognition performance of the proposed approach (Ours) in comparison with two previous works (SLF \cite{Srinivas_2019_CVPR_Workshops} and FTC \cite{DBLP:conf/iwbf/TerhorstD0K20}). The FNMR is shown at different FMR thresholds. Base refers to the unmodified FaceNet and VGGFace performance.
Even while making the recognition process more fair, in contrast to previous work, our approach consistently improves the global recognition performance.}
\label{tab:GlobalPerformance}
\begin{tabular}{llrrrrrrrrrrrr}
\Xhline{2\arrayrulewidth} 
               &              & \multicolumn{6}{c}{FaceNet}                                                              & \multicolumn{6}{c}{VGGFace}                                                              \\
               \cmidrule(lr){3-8} \cmidrule(lr){9-14}
               &    & \multicolumn{2}{c}{Adience} & \multicolumn{2}{c}{ColorFeret} & \multicolumn{2}{c}{Morph} & \multicolumn{2}{c}{Adience} & \multicolumn{2}{c}{ColorFeret} & \multicolumn{2}{c}{Morph} \\
               \hline
\parbox[t]{2mm}{\multirow{4}{*}{\rotatebox[origin=c]{90}{$10^{-3}$ FMR}}}  & Unnormalized & 0.4481 &         & 0.1460 &         & 0.0062 &         & 0.5201 &         & 0.2107 &          & 0.0465 &         \\
               & SLF \cite{Srinivas_2019_CVPR_Workshops}   & 0.4438 & 1.0\%   & 0.1229 & 15.8\%  & 0.0095 & -53.8\% & 0.4438 & 14.7\%  & 0.1229 & 41.7\%   & 0.0095 & 79.5\%  \\
               & FTC \cite{DBLP:conf/iwbf/TerhorstD0K20}    & 0.7109 & -58.6\% & 0.1406 & 3.7\%   & 0.0081 & -30.3\% & 0.7579 & -45.7\% & 0.4941 & -134.5\% & 0.0681 & -46.6\% \\
               & Ours         & 0.2694 & 39.9\%  & 0.1343 & 8.0\%   & 0.0029 & 53.2\%  & 0.4430 & 14.8\%  & 0.2203 & -4.6\%   & 0.0363 & 21.9\%  \\
\parbox[t]{2mm}{\multirow{4}{*}{\rotatebox[origin=c]{90}{$10^{-4}$ FMR}}} & Unnormalized & 0.7651 &         & 0.3299 &         & 0.0219 &         & 0.7404 &         & 0.3635 &          & 0.1180 &         \\
               & SLF \cite{Srinivas_2019_CVPR_Workshops}   & 0.6840 & 10.6\%  & 0.2381 & 27.8\%  & 0.0318 & -45.4\% & 0.6840 & 7.6\%   & 0.2381 & 34.5\%   & 0.0318 & 73.0\%  \\
               & FTC \cite{DBLP:conf/iwbf/TerhorstD0K20}    & 0.9160 & -19.7\% & 0.3406 & -3.2\%  & 0.0285 & -30.1\% & 0.9780 & -32.1\% & 0.8225 & -126.3\% & 0.1809 & -53.3\% \\
               & Ours         & 0.4800 & 37.3\%  & 0.2517 & 23.7\%  & 0.0121 & 44.7\%  & 0.6281 & 15.2\%  & 0.3474 & 4.4\%    & 0.0987 & 16.4\%  \\
\parbox[t]{2mm}{\multirow{4}{*}{\rotatebox[origin=c]{90}{$10^{-5}$ FMR}}} & Unnormalized & 0.9324 &         & 0.5403 &         & 0.0576 &         & 0.8782 &         & 0.5804 &          & 0.2171 &         \\
               & SLF \cite{Srinivas_2019_CVPR_Workshops}  & 0.8074 & 13.4\%  & 0.3658 & 32.3\%  & 0.0768 & -33.3\% & 0.8074 & 8.1\%   & 0.3658 & 37.0\%   & 0.0768 & 64.6\%  \\
               & FTC \cite{DBLP:conf/iwbf/TerhorstD0K20}   & 0.9791 & -5.0\%  & 0.6009 & -11.2\% & 0.0743 & -28.9\% & 0.9976 & -13.6\% & 0.9765 & -68.3\%  & 0.3463 & -59.5\% \\
               & Ours         & 0.6813 & 26.9\%  & 0.3979 & 26.4\%  & 0.0371 & 35.6\%  & 0.7685 & 12.5\%  & 0.4778 & 17.7\%   & 0.0371 & 82.9\% \\
               \Xhline{2\arrayrulewidth} 
\end{tabular}
\end{table*}

\vspace{4mm}
\section{Conclusion}
\label{sec:Conclusion}










Despite the progress achieved by current face recognition systems, recent works showed that biometric systems impose a strong bias against subgroups of the population.
Consequently, there is an increased need for solutions that increase the fairness of such systems.
Previous works focused on learning bias-mitigated face representations.
However, these solutions are often hardly-integrable and degrade the overall recognition performance.
In this work, we propose a novel fair score normalization approach to mitigate bias from recognition systems. 
Our unsupervised score normalization approach is easily-integrable into existing systems and significantly enhances the system's overall recognition performance.
Integrating the idea of individual fairness, our solution aims at treating similar individuals similarly.
The experiments were conducted on three publicly available datasets captured under various conditions and on two kinds of face embeddings.
The results show that the proposed approach significantly reduces demographic-bias, e.g. it mitigates ethnic-bias by 17.4-32.8\%.
Additionally, it mitigates bias more consistently over demographic domains than related works and strongly enhances the overall recognition performance, e.g. by 16.4-82.90\% on the Morph benchmark.
In contrast to related works, our method jointly achieves the following points: it (a) does not need additional soft-biometric labels during training or inference time, (b) can be easily integrated into existing face recognition systems, (c) enhances the total face recognition performance, and (d) leads to a consistent bias-mitigation. Moreover, it is, by design, not limited to face biometrics.


\section*{Acknowledgments}
This research work has been funded by the German Federal Ministery of Education and Research and the Hessen State Ministry for Higher Education, Research and the Arts within their joint support of the National Research Center for Applied Cybersecurity. Portions of the research in this paper use the FERET database of facial images collected under the FERET program, sponsored by the Counterdrug Technology Development Program Office.

\bibliographystyle{ieee}
\bibliography{egbib}

%

\end{document}